\documentclass[runningheads]{llncs}
\usepackage[T1]{fontenc}
\usepackage{graphicx}
\usepackage{capt-of}  
\usepackage{cuted}
\usepackage[backend=biber, sorting=none, giveninits=true]{biblatex}
\usepackage{graphicx}%
\usepackage{multirow}%
\usepackage{amsmath,amssymb,amsfonts}%
\usepackage{mathrsfs}%
\usepackage[figuresright]{rotating}%
\usepackage{xcolor}%
\usepackage{textcomp}%
\usepackage{manyfoot}%
\usepackage{booktabs}%
\usepackage{algorithm}%
\usepackage{algorithmicx}%
\usepackage{algpseudocode}%
\usepackage{program}%
\usepackage{listings}%
\usepackage{wrapfig}
\usepackage{vruler}
\usepackage{setspace}
\usepackage{comment}
\usepackage{siunitx}
\usepackage{booktabs}

\usepackage{multicol}

\begin{document}
%
\title{The Optimal Choice of Hypothesis \\Is the Weakest, Not the Shortest}

%
%
\author{Michael Timothy Bennett\inst{1}\\\orcidID{0000-0001-6895-8782} 
}
\authorrunning{Michael Timothy Bennett}
%
\institute{The Australian National University
\\\email{michael.bennett@anu.edu.au}\\
\url{http://www.michaeltimothybennett.com/}
}
\maketitle              

\begin{abstract}
If $A$ and $B$ are sets such that $A \subset B$, generalisation may be understood as the inference from $A$ of a hypothesis sufficient to construct $B$. One might infer any number of hypotheses from $A$, yet only some of those may generalise to $B$. How can one know which are likely to generalise? One strategy is to choose the shortest, equating the ability to compress information with the ability to generalise (a ``proxy for intelligence”). We examine this in the context of a mathematical formalism of enactive cognition. We show that compression is neither necessary nor sufficient to maximise performance (measured in terms of the probability of a hypothesis generalising). We formulate a proxy unrelated to length or simplicity, called weakness. We show that if tasks are uniformly distributed, then there is no choice of proxy that performs at least as well as weakness maximisation in all tasks while performing strictly better in at least one. 
In experiments comparing maximum weakness and minimum description length in the context of binary arithmetic, the former generalised at between $1.1$ and $5$ times the rate of the latter. We argue this demonstrates that weakness is a far better proxy, and explains why Deepmind's Apperception Engine is able to generalise effectively\footnote{Appendices are to be found on GitHub \cite{bennett2023appendices}.}.

\end{abstract}

\section{Introduction}
\label{intro}
If $A$ and $B$ are sets such that $A \subset B$, generalisation may be understood as the inference from $A$ of a hypothesis sufficient to construct $B$. One might infer any number of hypotheses from $A$, yet only some of those may generalise to $B$. How can one know which are likely to generalise?
According to Ockham's Razor, the simpler of two explanations is the more likely \cite{sober2015}. Simplicity is not itself a measurable property, so the minimum description length principle \cite{rissanen1978} relates simplicity to length. Shorter representations are considered to be simpler, and tend to generalise more effectively. This is often applied in the context of induction by comparing the length of programs that explain what is observed (to chose the shortest, all else being equal). 
The ability to identify shorter representations is compression, and the ability to generalise is arguably intelligence \cite{chollet2019}. Hence the ability to compress information is often portrayed as a proxy for intelligence \cite{chaitin2006}, even serving as the foundation \cite{solomonoff_1964a,solomonoff_1964b,kolmogorov_1963} of the theoretical super-intelligence AIXI \cite{hutter2010}.
That compression is a good proxy seems to have gone unchallenged. The optimal choice of hypothesis is widely considered to be the shortest. We show that it is not\footnote{This proof is conditional upon certain assumptions regarding the nature of cognition as enactive, and a formalism thereof.}. We present an alternative, unrelated to description length, called weakness. We prove that to maximise the probability that one's hypotheses generalise, it is necessary and sufficient to infer the weakest valid hypotheses possible\footnote{Assuming tasks are uniformly distributed, and weakness is well defined.}.

\section{Background definitions}
\label{definitions}

To do so, we employ a formalism of enactive cognition \cite{bennett2022a,bennett2023a,bennett2023c,bennett2023d,ward2017,bennett2023appendices}, in which sets of declarative programs are related to one another in such a way as to form a lattice. This unusual representation is necessary to ensure that both the weakness and description length of a hypothesis are well defined\footnote{An example of how one might translate propositional logic into this representation is given at the end of this paper. It is worth noting that this representation of logical formulae addresses the symbol grounding problem \cite{harnad1990}, and was specifically constructed to address subjective performance claims in the context of AIXI \cite{leike2015}.}. This formalism can be understood in three steps. 
\begin{enumerate} {
    \item The environment is represented as a set of declarative programs. 
    \item A finite subset of the environment is used to define a language with which to write statements that behave as logical formulae.
    \item Finally, induction is formalised in terms of tasks made up of these statements. }
\end{enumerate}

\begin{definition}[environment]\label{d1}
\begin{itemize}{
    \item  We assume a set $\Phi$ whose elements we call \textbf{states}, one of which we single out as the \textbf{present state}\footnote{Each state is just reality from the perspective of a point along one or more dimensions. States of reality must be separated by something, or there would be only one state of reality. For example two different states of reality may be reality from the perspective of two different points in time, or in space and so on.}.
    \item A \textbf{declarative program} is a function $f : \Phi \rightarrow \{true, false\}$, and we write $P$ for the set of all declarative programs. By an \textbf{objective truth} about a state $\phi$, we mean a declarative program $f$ such that $f(\phi) = true$.
    }
\end{itemize}
\end{definition}

\begin{definition}[implementable language] \label{d2}
\begin{itemize}{
    \item $\mathfrak{V} = \{V \subset P : V \ is \ finite\}$ is a set whose elements we call \textbf{vocabularies}, one of which we single out as \textbf{the vocabulary} $\mathfrak{v}$ for an implementable language.
    \item ${L_\mathfrak{v}} = \{ l \subseteq \mathfrak{v} : \exists \phi \in \Phi \ (\forall p \in l : p(\phi) = true) \}$ is a set whose elements we call \textbf{statements}\footnote{Statements are the logical formulae about which we will reason. }. $L_\mathfrak{v}$ follows from $\Phi$ and $\mathfrak{v}$. We call $L_\mathfrak{v}$ an \textbf{implementable language}.
    \item  $l \in {L_\mathfrak{v}}$ is \textbf{true} iff the present state is $\phi$ and $\forall p \in l : p(\phi) = true$.
    \item The \textbf{extension of a statement} $a \in {L_\mathfrak{v}}$ is $Z_a = \{b \in {L_\mathfrak{v}} : a \subseteq b\}$.
    \item The \textbf{extension of a set of statements} $A \subseteq {L_\mathfrak{v}}$ is $Z_A = \bigcup\limits_{a \in A} Z_a$.
    }
\end{itemize}

\noindent {\normalfont(Notation)} $Z$ with a subscript is the extension of the subscript\footnote{e.g. $Z_s$ is the extension of $s$.}. Lower case letters represent statements, and upper case represent sets of statements.

\end{definition}

\begin{definition}[{$\mathfrak{v}$}-task]\label{d3} For a chosen $\mathfrak{v}$, a task $\alpha$ is $\langle {S}_\alpha, {D}_\alpha, {M}_\alpha \rangle$ where:\begin{itemize}{ 
    \item ${S}_\alpha \subset L_\mathfrak{v}$ is a set whose elements we call \textbf{situations} of $\alpha$.
    \item ${S_\alpha}$ has the extension $Z_{S_\alpha}$, whose elements we call \textbf{decisions} of $\alpha$. 
    \item ${D_\alpha} = \{z \in Z_{S_\alpha} : z \ is \ correct \}$ is the set of all decisions which complete $\alpha$. 
    \item ${M_\alpha} = \{l \in L_\mathfrak{v} : {Z}_{S_\alpha} \cap Z_{l} = {D_\alpha}\}$ whose elements we call \textbf{models} of $\alpha$.}
\end{itemize}
$\Gamma_\mathfrak{v}$ is the set of all tasks\footnote{For example, we might represent chess as a supervised learning problem where $s \in S_\alpha$ is the state of a chessboard, $z \in Z_s$ is a sequence of moves by two players that begins in $s$, and $d \in D_\alpha \cap Z_s$ is such a sequence of moves that terminates in victory for one player in particular (the one undertaking the task).}.\\

\noindent{\normalfont(Notation)} If $\omega \in \Gamma_\mathfrak{v}$, then we will use subscript $\omega$ to signify parts of $\omega$, meaning one should assume $\omega = \langle {S}_\omega, {D}_\omega, {M}_\omega \rangle$ even if that isn't written.\\

\noindent {\normalfont(How a task is completed)} Assume we've a $\mathfrak{v}$-task $\omega$ and a hypothesis $\textbf{h} \in L_\mathfrak{v}$ s.t.\begin{enumerate}{
    \item we are presented with a situation ${s} \in {S}_\omega$, and
    \item we must select a decision $z \in Z_{s} \cap Z_\textbf{h}$.
    \item If $z \in {D}_\omega$, then $z$ is correct and the task is complete. This occurs if $\textbf{h} \in {M}_\omega$.}
\end{enumerate} 
\end{definition}

\section{Formalising induction}
\label{learning}

\begin{definition}[probability]\label{d4} We assume a uniform distribution over $\Gamma_\mathfrak{v}$.
\end{definition}

\begin{definition}[generalisation]\label{d5} A statement $l$ generalises to $\alpha \in \Gamma_\mathfrak{v}$ iff $l \in M_\alpha$. We say $l$ generalises from $\alpha$ to  $\mathfrak{v}$-task $\omega$ if we first obtain ${l}$ from ${M}_\alpha$ and then find it generalises to $\omega$.
\end{definition}

\begin{definition}[child and parent]\label{d6} A $\mathfrak{v}$-task $\alpha$ is a child of $\mathfrak{v}$-task $\omega$ if ${S}_\alpha \subset {S}_\omega$ and ${D}_\alpha \subseteq {D}_\omega$. 
This is written as $\alpha \sqsubset \omega$. If $\alpha \sqsubset \omega$ then $\omega$ is then a parent of $\alpha$.
\end{definition}

A proxy is meant to estimate one thing by measuring another. In this case, if intelligence is the ability to generalise \cite{bennett2022a,chollet2019}, then a greater proxy value is meant to indicate that a statement is more likely to generalise. Not all proxies are effective (most will be useless). We focus on two in particular.

\begin{definition}[proxy for intelligence]\label{d7} A proxy is a function parameterized by a choice of $\mathfrak{v}$ such that $q_\mathfrak{v} : L_\mathfrak{v} \rightarrow \mathbb{N}$. The set of all proxies is $Q$.\\

\noindent{\normalfont(Weakness)} The weakness of a statement $l$ is the cardinality of its extension $\lvert Z_{l} \rvert$. There exists $q_\mathfrak{v} \in Q$ such that $q_\mathfrak{v}(l) = \lvert Z_{l} \rvert$.\\

\noindent{\normalfont(Description length)} The description length of a statement $l$ is its cardinality $\lvert {l} \rvert$. Longer logical formulae are considered less likely to generalise \cite{rissanen1978}, and a proxy is something to be maximised, so description length as a proxy is $q_\mathfrak{v} \in Q$ such that $q_\mathfrak{v}(l) = \frac{1}{\lvert l \rvert}$.
\end{definition}

A child task may serve as an ostensive definition \cite{gupta2021} of its parent, meaning one can generalise from child to parent. 
\begin{definition}[induction]\label{d8} $\alpha$ and $\omega$ are $\mathfrak{v}$-tasks such that $\alpha \sqsubset \omega$. Assume we are given a proxy $q_\mathfrak{v} \in Q$, the complete definition of $\alpha$ and the knowledge that $\alpha \sqsubset \omega$. We are not given the definition of $\omega$. The process of induction would proceed as follows:
\begin{enumerate}{
    \item Obtain a hypothesis by computing a model $\mathbf{h} \in \underset{{m} \in {M}_\alpha}{\arg\max} \ q_\mathfrak{v}(m)$.
    \item If $\mathbf{h} \in {M}_\omega$, then we have generalised from $\alpha$ to $\omega$.}
\end{enumerate}
\end{definition}

\section{Proofs}
\label{proofs}

\begin{proposition}[sufficiency]\label{p1}Weakness is a proxy sufficient to maximise the probability that induction generalises from $\alpha$ to $\omega$.\end{proposition}
\begin{proof}
You're given the definition of $\mathfrak{v}$-task $\alpha$ from which you infer a hypothesis $\mathbf{h} \in {M}_\alpha$. $\mathfrak{v}$-task $\omega$ is a parent of $\alpha$ to which we wish to generalise:
\begin{enumerate} {
    \item The set of statements which \textit{might} be decisions addressing situations in ${S}_\omega$ and not ${S}_\alpha$, is $\overline{Z_{{S}_\alpha}} = \{ l \in L_\mathfrak{v} : l \notin Z_{{S}_\alpha} \}$.
    \item For any given $\mathbf{h} \in {M}_\alpha$, the extension $Z_\mathbf{h}$ of $\mathbf{h}$ is the set of decisions  $\mathbf{h}$ implies. The subset of $Z_\mathbf{h}$ which fall outside the scope of what is required for the known task $\alpha$ is $\overline{Z_{{S}_\alpha}} \cap Z_{\mathbf{h}}$ (because $Z_{{S}_\alpha}$ is the set of all decisions we might make when attempting $\alpha$, and so the set of all decisions that can't be made when undertaking $\alpha$ is $\overline{Z_{{S}_\alpha}}$ because those decisions occur in situations that aren't part of ${S}_\alpha$). 
    \item $\lvert \overline{Z_{{S}_\alpha}} \cap Z_{\mathbf{h}} \rvert$ increases monotonically with $\lvert Z_\mathbf{h} \rvert$, because $\forall z \in Z_m : z \notin \overline{Z_{{S}_\alpha} } \rightarrow z \in Z_{{S}_\alpha}$.
    \item $2^{\lvert \overline{Z_{{S}_\alpha}} \rvert}$ is the number of tasks which fall outside of what it is necessary for a model of $\alpha$ to generalise to (this is just the powerset of $\overline{Z_{{S}_\alpha}}$ defined in step 2), and $2^{\lvert \overline{Z_{{S}_\alpha}} \cap Z_{\mathbf{h}} \rvert}$ is the number of those tasks to which a given $\mathbf{h} \in M_\alpha$ does generalise.
    \item Therefore the probability that a given model $\mathbf{h} \in {M}_\alpha$ generalises to the unknown parent task $\omega$ is $$p(\mathbf{h} \in {M}_\omega \mid \mathbf{h} \in {M}_\alpha, \alpha \sqsubset \omega) = \frac{2^{\lvert \overline{Z_{{S}_\alpha}} \cap Z_{\mathbf{h}} \rvert}}{2^{\lvert \overline{Z_{{S}_\alpha}} \rvert}}$$}
\end{enumerate}
$p(\mathbf{h} \in {M}_\omega \mid \mathbf{h} \in {M}_\alpha, \alpha \sqsubset \omega)$ is maximised when $\lvert Z_\mathbf{h} \rvert$ is maximised. 
\end{proof}

\begin{proposition}[necessity]\label{p2}To maximise the probability that induction generalises from $\alpha$ to $\omega$, it is necessary to use weakness as a proxy, or a function thereof\footnote{For example we might use weakness multiplied by a constant to the same effect. }.\end{proposition}

\begin{proof} 
Let $\alpha$ and $\omega$ be defined exactly as they were in the proof of prop. \ref{p1}.
\begin{enumerate} {
    \item If $\mathbf{h} \in {M}_\alpha$ and $Z_{{S}_\omega} \cap Z_{\mathbf{h}} = {D}_\omega$, then it must be he case that ${D}_\omega \subseteq Z_{\mathbf{h}}$. 
    \item If $\lvert Z_{\mathbf{h}} \rvert < \lvert {D}_\omega \rvert$ then generalisation cannot occur, because that would mean that ${D}_\omega \not\subseteq Z_{\mathbf{h}}$. 
    \item Therefore generalisation is only possible if $\lvert Z_{{m}} \rvert \ge \lvert {D}_\omega \rvert$, meaning a sufficiently weak hypothesis is necessary to generalise from child to parent.
    \item The probability that $\lvert Z_{{m}} \rvert \ge \lvert {D}_\omega \rvert$ is maximised when $\lvert Z_{{m}} \rvert$ is maximised. Therefore to maximise the probability induction results in generalisation, it is necessary to select the weakest hypothesis. 
    }
\end{enumerate}
To select the weakest hypothesis, it is necessary to use weakness (or a function thereof) as a proxy.
\end{proof}

\begin{remark}[prior]\label{rm4} {\normalfont 
The above describes inference from a child to a parent. However, it follows that increasing the weakness of a statement increases the probability that it will generalise to any task (not just a parent of some given child). As tasks are uniformly distributed, every statement in $L_\mathfrak{v}$ is a model to one or more tasks, and the number of tasks to which each statement $l \in L_\mathfrak{v}$ generalises is $2^{\lvert Z_l\rvert}$. Hence the probability of generalisation\footnote{$\frac{2^{\lvert Z_\mathbf{h} \rvert}}{2^{\lvert L_\mathfrak{v} \rvert}}$ is maximised when $\mathbf{h} = \emptyset$, because the optimal hypothesis given no information is to assume nothing (you've no sequence to predict, so why make assertions that might contradict the environment?).} to $\omega$ is $p(\mathbf{h} \in M_\omega \mid \mathbf{h} \in {L}_\mathfrak{v}) = \frac{2^{\lvert Z_\mathbf{h} \rvert}}{2^{\lvert L_\mathfrak{v} \rvert}}$. This assigns a probability to every statement $l \in L_\mathfrak{v}$ given an implementable language. It is a probability distribution in the sense that the probability of mutually exclusive statements sums to one\footnote{Two statements $a$ and $b$ are mutually exclusive if $a \not\in Z_b$ and $b \not\in Z_a$, which we'll write as $\mu(a,b)$. Given $x \in L_\mathfrak{v}$, the set of all mutually exclusive statements is a set $K_x \subset L_\mathfrak{v}$ such that $x \in K_x$ and $\forall a, b \in K_x : \mu(a,b)$. It follows that $\forall x \in L_\mathfrak{v}, \underset{b \in K_x}{\sum} p(b) = 1$.}. This prior may be considered universal in the very limited sense that it assigns a probability to every conceivable hypothesis (where what is conceivable depends upon the implementable language) absent any parameters or specific assumptions about the task as with AIXI's intelligence order relation \cite[def. 5.14 pp. 147]{hutter2010}\footnote{We acknowledge that some may object to the term universal, because $\mathfrak{v}$ is finite.}. As the vocabulary $\mathfrak{v}$ is finite, $L_\mathfrak{v}$ must also be finite, and so $p$ is computable.}
\end{remark}
We have shown that, if tasks are uniformly distributed, then weakness is a necessary and sufficient proxy to maximise the probability that induction generalises. It is important to note that another proxy may perform better given cherry-picked combinations of child and parent task for which that proxy is suitable. However, such a proxy would necessarily perform worse given the uniform distribution of all tasks. Can the same be said of description length?

\begin{proposition}\label{p4} Description length is neither a necessary nor sufficient proxy for the purposes of maximising the probability that induction generalises.
\end{proposition}

\begin{proof}
In propositions \ref{p1} and \ref{p2} we proved that weakness is a necessary and sufficient choice of proxy to maximise the probability of generalisation. 
It follows that either maximising $\frac{1}{\lvert m \rvert}$ (minimising description length) maximises $\lvert Z_{m} \rvert$ (weakness), or minimisation of description length is unnecessary to maximise the probability of generalisation. Assume the former, and we'll construct a counterexample with $\mathfrak{v} = \{a,b,c,d,e,f,g,h,j,k,z \}$ s.t.
$L_\mathfrak{v} = \{   
    \{a,b,c,d,j,k,z\}, 
    \{e,b,c,d,k\},\\
    \{a,f,c,d,j\},   
    \{e,b,g,d,j,k,z\},  
    \{a,f,c,h,j,k\},
    \{e,f,g,h,j,k\} 
    \}$ 
and a task $\alpha$ where
\begin{itemize}{
    \item ${S}_\alpha = \{   \{a,b\}, \{e,b\}
    \}$
    \item ${D}_\alpha = \{   \{a,b,c,d,j,k,z\}, \{e,b,g,d,j,k,z\}
    \}$
    \item ${M}_\alpha = \{\{z\}, \{j, k \} \}$}
\end{itemize}
Weakness as a proxy selects $\{j,k\}$, while description length as a proxy selects $\{z\}$. This demonstrates the minimising description length does not necessarily maximise weakness, and maximising weakness does not minimise description length. As weakness is necessary and sufficient to maximise the probability of generalisation, it follows that minimising description length is neither.
\end{proof}

\section{Experiments}
\label{experiments}
Included with this paper is a Python script to perform two experiments using PyTorch with CUDA, SymPy and $A^*$ \cite{paszke2019,kirk2007,meurer2017,hart1968} (see technical appendix for details). 
In these two experiments, a toy program computes models to 8-bit string prediction tasks (binary addition and multiplication). The purpose of these experiments was to compare weakness and description length as proxies. 

\subsection{Setup} To specify tasks with which the experiments would be conducted, we needed a vocabulary $\mathfrak{v}$ with which to describe simple 8-bit string prediction problems. There were 256 states in $\Phi$, one for every possible 8-bit string. The possible statements were then all the expressions regarding those $8$ bits that could be written in propositional logic (the simple connectives $\lnot$, $\land$ and $\lor$ needed to perform binary arithmetic -- a written example of how propositional logic can be used in to specify $\mathfrak{v}$ is also included in the appendix). In other words, for each statement in $L_\mathfrak{v}$ there existed an equivalent expression in propositional logic. For efficiency, these statements were implemented as either PyTorch tensors or SymPy expressions in different parts of the program, and converted back and forth as needed (basic set and logical operations on these propositional tensor representations were implemented for the same reason). A $\mathfrak{v}$-task was specified by choosing ${D}_n  \subset L_\mathfrak{v}$ such that all ${d} \in {D}_n$ conformed to the rules of either binary addition or multiplication with 4-bits of input, followed by 4-bits of output. 

\subsection{Trials}
Each experiment had parameters were ``operation'' and ``number\_of\_trials''. For each trial the number $\lvert {D}_k \rvert$ of examples ranged from $4$ to $14$.
A trial had $2$ phases. 
\subsubsection{Training phase:}
\begin{enumerate} {
    \item A task $n$ (referred to in code as ${T}_n$) was generated: 
    \begin{enumerate} {
        \item First, every possible 4-bit input for the chosen binary operation was used to generate an 8-bit string. These 16 strings then formed ${D}_n$.
        \item A bit between 0 and 7 was then chosen, and ${S}_n$ created by cloning ${D}_n$ and deleting the chosen bit from every string (${S}_n$ contained 16 different 7-bit strings, each of which was a sub-string of an element of ${D}_n$).}
    \end{enumerate} 
    \item A child-task $k =  \langle {S}_k, {D}_k, M_k  \rangle $ (referred to in code as ${T}_k$) was sampled (assuming a uniform distribution over children) from the parent task ${T}_n$. Recall, $\lvert {D}_k \rvert$ was determined as a parameter of the trial.
    \item From ${T}_k$ two models were then generated; a weakest $c_w$, and a MDL $c_{mdl}$. }
\end{enumerate}

\subsubsection{Testing phase:}
For each model $c \in \{c_w, c_{mdl}\}$, the testing phase was as follows:
\begin{enumerate} {
    \item The extension $Z_c$ of $c$ was then generated.
    \item A prediction ${D}_{recon}$ was made s.t. ${D}_{recon} = \{z \in Z_c : \exists {s} \in {S}_n \ ({s} \subset z ) \}$.
    \item ${D}_{recon}$ was then compared to the ground truth ${D}_n$, and results recorded. }
\end{enumerate}
Between $75$ and $256$ trials were run for each value of the parameter $\lvert {D}_k \rvert$. Fewer trials were run for larger values of $\lvert {D}_k \rvert$ as these took longer to process. The results of these trails were then averaged for each value of $\lvert {D}_k \rvert$.

\subsection{Results}
Two sorts of measurements were taken for each trial.
The first was \textbf{the rate at generalisation occurred}. Generalisation was deemed to have occurred where ${D}_{recon} = {D}_n$. The number of trials in which generalisation occurred was measured, and divided by $n$ to obtain the rate of generalisation for $c_w$ and $c_{mdl}$. Error was computed as a Wald 95$\%$ confidence interval.
The second measurement was \textbf{the average extent to which models generalised}. Even where ${D}_{recon} \neq {D}_n$, the extent to which models generalised could be ascertained. $\frac{\lvert {D}_{recon} \cap {D}_n\rvert}{\lvert {D}_n\rvert}$ was measured and averaged for each value of $\lvert {D}_k\rvert$, and the standard error computed. The results (see tables \ref{table1} and \ref{table2}) demonstrate that weakness is a better proxy for intelligence than description length.  
The generalisation rate for $c_w$ was between $110-500\%$ of $c_{mdl}$, and the extent was between $103-156\%$. 

\begin{table}
  \caption{Results for Binary Addition}
  \label{table1}
  \centering
  \begin{tabular}{lllllllll}
    \toprule
    \multicolumn{1}{l}{}
    & $c_w$ & & & & $c_{mdl}$ \\
    \cmidrule(r){2-3}
    \cmidrule(r){4-5}
    \cmidrule(r){6-7}
    \cmidrule(r){8-9}
    $\lvert {D}_k \rvert$     & Rate & $\pm 95 \%$    & AvgExt & StdErr & Rate & $\pm 95 \%$       & AvgExt & StdErr \\
    \midrule
    6 & .11 & .039 & .75 & .008 & .10 & .037 & .48 & .012    \\
    10 & .27 & .064 & .91 & .006  & .13 & .048 & .69 & .009     \\
    14 & .68 & .106 & .98 & .005  & .24 & .097 & .91 & .006  \\
    \bottomrule
  \end{tabular}
\end{table}
\begin{table}
  \caption{Results for  Binary Multiplication}
  \label{table2}
  \centering
  \begin{tabular}{lllllllll}
    \toprule
    \multicolumn{1}{l}{}
    & $c_w$ & & & & $c_{mdl}$ \\
    \cmidrule(r){2-3}
    \cmidrule(r){4-5}
    \cmidrule(r){6-7}
    \cmidrule(r){8-9}
    $\lvert {D}_k \rvert$     & Rate & $\pm 95 \%$    & AvgExt & StdErr & Rate & $\pm 95 \%$       & AvgExt & StdErr \\
    \midrule
    6 & .05  & .026 & .74 & .009 & .01  & .011 & .58 & .011    \\
    10 & .16 & .045 & .86 & .006  & .08 & .034 & .78 & .008     \\
    14 & .46 & .061 & .96 & .003  & .21 & .050 & .93 & .003  \\
    \bottomrule
  \end{tabular}
\end{table}

\section{Concluding remarks}
\label{conclusion}

We have shown that, if tasks are uniformly distributed, then weakness maximisation is necessary and sufficient to maximise the probability that induction will produce a hypothesis that generalises. It follows that there is no choice of proxy that performs at least as well as weakness maximisation across all possible combinations of child and parent task while performing strictly better in at least one. 
We've also shown that the minimisation of description length is neither necessary nor sufficient. This calls into question the relationship between compression and intelligence
\cite{chaitin2006,orallo2010,legg2011}, at least in the context of enactive cognition. This is supported by our experimental results, which demonstrate that weakness is a far better predictor of whether a hypothesis will generalise, than description length. Weakness should not be conflated with Ockham's Razor. A simple statement need not be weak, for example ``all things are blue crabs''. Likewise, a complex utterance can assert nothing. Weakness is a consequence of extension, not form. If weakness is to be understood as an epistemological razor, it is this (which we humbly suggest naming ``Bennett's Razor''): $$\textit{Explanations should be no more specific than necessary.}\footnote{We do not know which possibilities will eventuate. A less specific statement contradicts fewer possibilities. Of all hypotheses sufficient to explain what we perceive, the least specific is most likely. }$$

\subsubsection{The Apperception Engine:}
\label{bias}
The Apperception Engine \cite{evans_2020a,evans_2020b,evans_2021b} (Evans et. al. of Deepmind) is an inference engine that generates hypotheses that generalise often. To achieve this, Evans formalised Kant's philosophy to give the engine a ``strong inductive bias''. The engine forms hypotheses from only very general assertions, meaning logical formulae which are universally quantified. That is possible because the engine uses language specifically tailored to efficiently represent the sort of sequences to which it is applied. 
Our results suggest a simpler and more general explanation of why the engine's hypotheses generalise so well. The tailoring of logical formulae to represent certain sequences amounts to a choice of $\mathfrak{v}$, and the use of only universally quantified logical formulae ensures the resulting hypothesis is weak. 
Obviously this can work well, but only for the subset of possible tasks that the vocabulary is able to describe in this way (anything else will not be able to be represented as a universally quantified rule, and so will not be represented at all \cite{bennett2022b}). This illustrates how future research
may explore choices of $\mathfrak{v}$ in aid of more efficient induction in particular sorts of task, such as the inference of linguistic meaning and intent (see appendix).

\subsubsection{Neural networks:}
How might a task be represented in the context of a function? Though we use continuous real values in base $10$ to formalise neural networks, all computation still takes place in a discrete, finite and binary system. A finite number of imperative programs composed a finite number of times may be represented by a finite set of declarative programs. 
Likewise, activations within a network given an input can be represented as a finite set of declarative programs, expressing a decision. The choice of architecture specifies the vocabulary in which this is written, determining what sort of relations can be described according to the Chomsky Hierarchy \cite{deletang2022}. The reason why LLMs are so prone to fabrication and inconsistency may be because they are optimised only to minimise loss, rather than maximise weakness \cite{bennett2022a}. Perhaps grokking \cite{power2022} can be induced by optimising for weakness. Future research should investigate means by which weakness can be maximised in the context of neural networks.

\printbibliography

\end{document}